\documentclass[12pt]{article}
\usepackage[utf8]{inputenc}
\usepackage{graphicx}
\usepackage{caption}
\usepackage{subcaption}
\usepackage{amssymb, amsmath, amsfonts}
\usepackage{multicol}
\usepackage{algorithm}
\usepackage{authblk}

\usepackage{algpseudocode}

\DeclareMathOperator*{\argmax}{arg\,max}
\DeclareMathOperator*{\argmin}{arg\,min}

\algnewcommand\algorithmicinput{\textbf{Inputs:}}
\algnewcommand\INPUT{\item[\algorithmicinput]}

\algnewcommand\algorithmicoutput{\textbf{Outputs:}}
\algnewcommand\OUTPUT{\item[\algorithmicoutput]}

\graphicspath{{./figures/}}
\title{Kullback-Leibler Penalized Sparse Discriminant Analysis for Event-Related Potential Classification}

\author[1]{Victoria Peterson\thanks{vpeterson@sinc.unl.edu.ar}}
\author[1,2]{Hugo Leonardo Rufiner}
\author[3]{Ruben Daniel Spies}

\affil[1]{Instituto de Investigación en Señales, Sistemas e Inteligencia Computacional, UNL, CONICET, FICH, Ruta Nac. 168, km 472.4, 3000, Santa Fe, Argentina.}
\affil[2]{Facultad de Ingenier\'{\i}a, Universidad Nacional de Entre R\'{\i}os, Ruta Prov. 11, km 10, 3100, Oro Verde, Argentina.}
\affil[3]{Instituto de Matemática Aplicada del Litoral, UNL, CONICET, FIQ, 
Predio Dr. Alberto Cassano del CCT-CONICET-Santa Fe, Ruta Nac. 168, km 0, 3000, Santa Fe, Argentina.}

\begin{document}
\maketitle
\begin{abstract}
A brain computer interface (BCI) is a system which provides direct communication between the mind  of a person and the outside world by using only brain activity (EEG). The event-related potential (ERP)-based BCI problem consists of a binary pattern recognition. Linear discriminant analysis (LDA) is widely used to solve this type of classification problems, but it fails when the number of features is large relative to the number of observations. In this work we propose a penalized version of the sparse discriminant analysis (SDA), called Kullback-Leibler penalized sparse discriminant analysis (KLSDA). This method inherits both the discriminative feature selection and classification properties of SDA and it also improves SDA performance through the addition of Kullback-Leibler class discrepancy information. The KLSDA method is design to automatically select the optimal regularization parameters. Numerical experiments with two real ERP-EEG datasets show that this new method outperforms standard SDA. 
\end{abstract}


%\begin{keyword}
%\vspace{0.1in}
%
%Brain-Computer Interface \sep Event-Related Potential \sep Kullback-Leibler Divergence \sep Penalization \sep Sparse Discriminant Analysis
%%% keywords here, in the form: keyword \sep keyword


%\end{keyword}
\section{Introduction}
\label{Intro}
A brain computer interface (BCI) is a system that measures brain activity and converts it into an artificial output which is able to replace, restore or improve any normal output (neuromuscular or hormonal) used by a person to communicate and control his/her external or internal environment. Thus, BCI can significantly improve the quality of life of people with severe neuromuscular disabilities \cite{BCIcommunication}.

Communication between the brain of a person and the outside world can be appropriately established by means of a BCI system based on event-related potentials (ERPs), which are manifestations of neural activity as a consequence of certain infrequent or relevant stimuli. The main reason for using ERP-based BCI are: it is non-invasive, it requires minimal user training and it is quite robust (in the sense that it can be use by more than 90 \% of people) \cite{wolpawBCI}. One of the main components of such ERPs is the P300 wave, which is a positive deflection occurring in the scalp-recorded EEG approximately 300 ms after the stimulus has been applied. The P300 wave is unconsciously generated and its latency and amplitude vary between different EEG records of the same person, and even more, between EEG records of different persons \cite{Electroph}. By using the ``oddball'' paradigm \cite{Don-Far} an ERP-based BCI can decode desired commands from the subject by detecting those ERPs in the background EEG. In other words, a ERP-based BCI deals with a pattern recognition problem where two classes are involved: EEG with ERP (target class) and EEG without ERP (non-target class). 
%Figure \ref{ERP} shows a typical ERP signal estimated by synchronous averaging target EEG records. 
%\begin{figure}
%\centerline{
%\includegraphics[width=10cm]{erp}}
%\caption{Estimated ERP signal by synchronous averaging of EEG target segments.}
%\label{ERP}
%\end{figure}

Several authors have investigated in regards to which classification technique performs best for separating target from non-target EEG records \cite{Lottereview, Blankertz-tutorial, AdvanceChallenges, gareis2011}. All of these studies have concluded that linear discriminant analysis (LDA) is a very good classification scheme, resulting most of the times in the optimal performance while keeping the solution simple. It is timely to mention here that in a BCI classification scheme there are two main problems: the curse-of-dimensionality and the bias-variance trade-off \cite{Lottereview}. While the former is a consequence of working with a concatenation of multiple time points from multiple channels, the latter depicts the generalization capability of the classifier. 

However, it has been shown that LDA performs poorly when high dimensional data with small training sample is used. Different regularized solutions have been proposed \cite{krusienskicomparison, meanshrinkage, aggregation} and all of them have demonstrated that a regularized version of LDA can substantially improve the classification performance of the standard LDA. However, there is still much room for further improvement. In this regards, in the present work, we propose to achieve significant gain by building up a model which takes into account the following issues, which, to our knowledge, have not been jointly considered in any of the previous works:
\begin{itemize}
\item Solution sparsity needs to be kept as high as possible in order to maximize robustness while achieving good classification performance. It is timely to remember that the generalizability of a model tends to decrease as the sparsity decreases, i.e., as the number of parameters in the model increases \cite{bookstats}.
\item Since randomly selecting one variable out of a group of correlated variables may produce neglecting important information, groups of ``similar'' variables should be allowed into the model \cite{enet}.  
\item Model should allow for the inclusion of discriminant a priori information since that could greatly enhance separability between classes. 
\end{itemize}

With the above in mind, in this work we develop a penalized version of the sparse discriminant analysis (SDA) \cite{SLDA}, which we call Kullback-Leibler Penalized sparse discriminant analysis (KLSDA), with the main objective of solving the binary ERP classification problem. As far as we know SDA has never been used before in ERP-based BCI classification problems. Also, we shall compare our penalized version with the standard SDA in this context.

The organization of this article is as follows. In Section \ref{reviewLDA} we made a brief review on discriminant analysis. Our proposed new approach is presented in its full general formulation in Section \ref{ourKLSDA}. In Section \ref{datasets} the ERP-EEG databases used in the experiments are described. In Section \ref{Experiments} the experiments and results of the ERP-based EEG classification problem solved by KLSDA are shown. Finally, concluding remarks and future works are presented in Section \ref{Discussion}.
\section{Discriminant Analysis: a brief review}
\label{reviewLDA}
The LDA criterion is a well-known dimensionality reduction tool in the context of supervised classification. Its popularity is mainly due to its simplicity and robustness which lead to very high classification performances in many applications \cite{Duda}. 

Let $\mathbf{W}_{\text{1}},\dots, \mathbf{W}_{\text{K}}$ be $p$-dimensional random vectors whose distributions uniquely characterize each one of the $K$ classes of a given classification problem. In addition, let $\mathbf{X}$ be an ${n \times p }$ data matrix such that each row $\mathbf{x}_i$ is a realization of one and only one of the aforementioned random vectors, and let $\mathbf{z} \in \mathbb{R}^n$ be a categorical variable accounting for class membership; i.e., such that if pattern $\mathbf{x}_i$ is a realization of $\mathbf{W}_{\text{k}}$, then $z_i=k$. 

%In classification problems the main objective is to efficiently decide the class to which each pattern (observation $\mathbf{x}_i$) belongs. We assume that we are provided with a set of patterns of known class $\left\lbrace (\mathbf{x}_i, z_i), i=1, \dots, n \right\rbrace$, called the ``training set'', which is used for designing a classifier. To evaluate the classifier's performance we are also provided with a set of patterns of unknown class, called the ``testing set''.
The LDA method consists of finding $q < K$ discriminant vectors (directions), $\boldsymbol{\beta}_1, \dots, \boldsymbol{\beta}_q$ such that by projecting the data matrix $\mathbf{X}$ over those directions, the ``classes'' will be well separated one from each other. It is assumed that the random
vectors $\mathbf{W}_{\text{1}},\dots, \mathbf{W}_{\text{K}}$ are independently and normally distributed with a common covariance matrix $\boldsymbol{\Sigma}_t$. The procedure for finding the vectors $\boldsymbol{\beta}_j$ requires estimates of the within-class, the between-class and the total covariance matrices, $\boldsymbol{\Sigma}_w$, $\boldsymbol{\Sigma}_b$ and $\boldsymbol{\Sigma}_t$, respectively. These estimates are given by:
\[
\hat{\boldsymbol{\Sigma}}_w= \frac{1}{n}\sum_{k=1}^K \sum_{i \in I_k} (\mathbf{x}_i-\boldsymbol{\mu}_k)(\mathbf{x}_i-\boldsymbol{\mu}_k)^T,
\]
\[\hat{\boldsymbol{\Sigma}}_b= \frac{1}{n}\sum_{k=1}^K n_k(\boldsymbol{\mu}_k-\boldsymbol{\mu})(\boldsymbol{\mu}_k-\boldsymbol{\mu})^T,
\]
\[
\hat{\boldsymbol{\Sigma}}_t= \frac{1}{n}\sum_{i=1}^n(\mathbf{x}_i-\boldsymbol{\mu})(\mathbf{x}_i-\boldsymbol{\mu})^T,
\]
where $I_k$ and $n_k$ are the set of indices and the number of patterns belonging to class $k$, respectively, $\boldsymbol{\mu}_k \dot = $ $\frac{1}{n_k}\sum_{i \in I_k}\mathbf{x}_i$ is the class $k$ sample mean and $\boldsymbol{\mu} \dot = \frac{1}{n}\sum_{k=1}^{K}\boldsymbol{\mu}_k$ is the common sample mean. Note that $\hat{\boldsymbol{\Sigma}}_t=\hat{\boldsymbol{\Sigma}}_w + \hat{\boldsymbol{\Sigma}}_b$.

Let $\mathbf{B}$ be a $p \times q$ matrix whose $j^{th}$ columns is $\boldsymbol{\beta}_j$. The LDA method seeks to find the vectors $\boldsymbol{\beta}_j$ in such a way that they maximize separability between classes, which is achieved by simultaneously maximizing $\hat{\boldsymbol{\Sigma}}_b$ and minimizing $\hat{\boldsymbol{\Sigma}}_w$. Since $\hat{\boldsymbol{\Sigma}}_t=\hat{\boldsymbol{\Sigma}}_w + \hat{\boldsymbol{\Sigma}}_b$, this is equivalent to simultaneously maximizing $\hat{\boldsymbol{\Sigma}}_b$ and minimizing $\hat{\boldsymbol{\Sigma}}_t$, thus:
\begin{equation}
\mathbf{B}^*=\argmax_{\mathbf{B} \in \mathbb{R}^{p \times q}} (\mathbf{B}^T\hat{\boldsymbol{\Sigma}}_b\mathbf{B})(\mathbf{B}^T \hat{\boldsymbol{\Sigma}}_t\mathbf{B})^{-1}.
\label{LDA}
\end{equation}

Since the rank of $\hat{\boldsymbol{\Sigma}}_b$ is at most $K-1$, there are at most $K-1$ non-trivial solutions of problem (\ref{LDA}), and therefore there are at most $K-1$ discriminative vectors. Observe that, these vectors are precisely the directions along which the classes show maximum between-class covariance relative to their within-class covariance. Usually $q=K-1$.

It is known that LDA fails when the number of patterns is low in relation to the number of variables ($n << p$) \cite{penalized}. In this situation, the matrix $\hat{\boldsymbol{\Sigma}}_t$ becomes usually ill-conditioned, what produces a poor estimation of $\mathbf{B}$ and, as a consequence, a bad classification performance.

In the particular case $K=2$ (and therefore $q=1$), the solution to (\ref{LDA}) has the following explicit formulation:
\begin{equation}
\boldsymbol{\beta}^*= \hat{\boldsymbol{\Sigma}}_t^{-1}(\boldsymbol{\mu}_1-\boldsymbol{\mu}_2).
\label{FLDA}
\end{equation}
This special case is known as Fisher linear discriminant analysis (FLDA) \cite{fisher}. The FLDA approach can be formulated as a linear regression model \cite{fisher, Duda}. Let $\mathbf{X}$ be as before and let $\mathbf{y}$ be a $n$-dimensional vector such that $y_i=\frac{n_2}{n}$ or $y_i=-\frac{n_1}{n}$, depending on whether the $i^{th}$ observation belonging to class 1 or to class 2, respectively, and let us consider the following ordinary least squares problem (OLS):
\begin{equation}
\boldsymbol{\alpha}^*=\argmin_{\boldsymbol{\alpha} \in \mathbb{R}^p} \|\mathbf{y}-\boldsymbol{X\alpha}\|_2^2,
\label{OLS}
\end{equation}
whose solutions are all the vectors in the set $ \mathcal{N}(\mathbf{X}^T\mathbf{X})+(\mathbf{X}^T\mathbf{X})^{\dag}\mathbf{X}^T\mathbf{y}$, where ``$\dag$'' denotes the Moore-Penrose generalized inverse and $ \mathcal{N}(\mathbf{
X}^T\mathbf{X})$ denotes the null space of $\mathbf{X}^T\mathbf{X}$. If $\mathbf{X}^T\mathbf{X}$ is invertible, then Eq. (\ref{OLS}) has a unique solution given by $
\boldsymbol{\alpha}^*=(\mathbf{X}^T\mathbf{X})^{-1}\mathbf{X}^T\mathbf{y}$. For convenience it is assumed that $\boldsymbol{\mu}=0$, and therefore $\mathbf{X}^T\mathbf{X}=n\boldsymbol{\Sigma}_t$ and $\mathbf{X}^T\mathbf{y}=\frac{n_1n_2}{n}(\boldsymbol{\mu}_1-\boldsymbol{\mu}_2)$. Hence $\boldsymbol{\alpha}^*= \frac{n_1n_2}{n^2} \boldsymbol{\beta}^*$, where $\boldsymbol{\beta}^*$ is given by (\ref{FLDA}). Since the proportionality constant $\frac{n_1n_2}{n^2}$ is irrelevant to the direction of the solution, this proves that OLS (Eq. (\ref{OLS})) is equivalent to the FLDA method (Eq. (\ref{FLDA})).

Several works (\cite{flexlda, LSLDA, bookstats}, to cite a few) have extended the above OLS-LDA formulation to multiclass problems. It has been shown that the LDA solution can be obtained from a multivariate regression fit. In particular, Hastie et al. in \cite{flexlda}, introduced a richer and more flexible classification scheme to LDA, called \emph{optimal scoring}, which we briefly describe below. 

Let $\mathbf{X}$ be as before and $\mathbf{Y}$ be a $n \times K$ matrix of binary variables such that $y_{ij}$ is an indicator variable of whether the $i^{th}$ observation belongs to the $j^{th}$ class. Let us define $\boldsymbol{\Theta}=\left[\boldsymbol{\theta}_1, \dots, \boldsymbol{\theta}_q \right] \in \mathbb{R}^{K \times q}$, where the vectors $\boldsymbol{\theta}_j$ are recursively obtained, for $j=1,2, \dots, q$, as the solution of the following constrained least squares problem which resumes the optimal scoring:
\begin{align}
\nonumber
\left(\boldsymbol{\beta}_j, \boldsymbol{\theta}_j\right)&=\argmin_{\boldsymbol{\beta}\in \mathbb{R}^p, \boldsymbol{\theta}\in \mathbb{R}^K} \| \mathbf{Y}\boldsymbol{\theta}- \mathbf{X}\boldsymbol{\beta} \|_2^2, \\
&s.t. \; \frac{1}{n}\boldsymbol{\theta}^T\mathbf{Y}^T\mathbf{Y}\boldsymbol{\theta}=1, \; \boldsymbol{\theta}^T\mathbf{Y}^T\mathbf{Y}\boldsymbol{\theta}_l=0 \quad \forall l=1,2, \dots, j-1.
\label{optimalscoring}
\end{align}

Note that when $j=1$ the orthogonality condition in Eq. (\ref{optimalscoring}), which is imposed to avoid trivial solutions, is vacuous and hence it is not enforced. Details about the computational implementation to solve Eq. (\ref{optimalscoring}) can be found in \cite{flexlda}.

In the sequel we shall refer to $\boldsymbol{\theta}_j$ as the ``score vector''. Observe that $\boldsymbol{\theta}_j$ is the vector in $\mathbb{R}^K$ for which the mapping from $\mathbb{R}^{n \times K}$ to $\mathbb{R}^{n}$ defined by $\mathbf{Y} \rightarrow \mathbf{Y}\boldsymbol{\theta}_j$, results optimal for the constrained least squares problem defined by (\ref{optimalscoring}). This mapping is precisely what introduces more flexibility into the LDA framework since it transforms binary variables into real ones.

Clemmensen et al. \cite{SLDA} introduced a regularized version of the optimal scoring problem by adding two penalization terms to the functional in (\ref{optimalscoring}). These penalization terms on one side induce sparsity and on the other side allow correlated variables into the solution. This regularized LDA formulation, named SDA, consists on recursively solving for $j=1, 2, \dots, q$, the following problem:
\begin{align}
\nonumber
\left(\boldsymbol{\beta}_j, \boldsymbol{\theta}_j\right)&=\argmin_{\boldsymbol{\beta}\in \mathbb{R}^p, \boldsymbol{\theta}\in \mathbb{R}^K} \{ \| \mathbf{Y}\boldsymbol{\theta}- \mathbf{X}\boldsymbol{\beta} \|_2^2 + \lambda_1 \| \boldsymbol{\beta}\|_1 + \lambda_2\|\boldsymbol{\beta}\|_2^2 \}, \\
& s.t. \quad \frac{1}{n}\boldsymbol{\theta}^T\mathbf{Y}^T\mathbf{Y}\boldsymbol{\theta}=1, \; \boldsymbol{\theta}^T\mathbf{Y}^T\mathbf{Y}\boldsymbol{\theta}_l=0 \quad \forall l=1,2, \dots, j-1,
\label{SDA}
\end{align}
where $\lambda_1$ and $\lambda_2$ are predefined positive constants, called regularization parameters, which balance the amount of sparsity and the correlation of variables, respectively. Later on we shall analyse appropriate forms for selecting those regularization parameters.
  
Problem (\ref{SDA}) is alternately and iteratively solved as follows. At first $\boldsymbol{\theta}_j$ is hold fixed and optimization is performed with respect to $\boldsymbol{\beta}_j$. Then $\boldsymbol{\beta}_j$ is hold fixed and optimization is performed with respect to $\boldsymbol{\theta}_j$. The following two steps are iterated:
\begin{enumerate}
\item[1.] For given (fixed) $\boldsymbol{\theta}_j$, solve:
\begin{equation}
\boldsymbol{\beta}_j= \argmin_{\boldsymbol{\beta} \in \mathbb{R}^p} \{ \| \mathbf{Y}\boldsymbol{\theta}_j- \mathbf{X}\boldsymbol{\beta} \|_2^2 +\lambda_1\|\boldsymbol{\beta}\|_1 + \lambda_2 \|\boldsymbol{\beta}\|_2^2 \}.
\label{enet}
\end{equation}
\item[2.] For given (fixed) $\boldsymbol{\beta}_j$, solve:
\begin{align*}
\boldsymbol{\theta}_j&=\argmin_{\boldsymbol{\theta} \in \mathbb{R}^K} \| \mathbf{Y}\boldsymbol{\theta}- \mathbf{X}\boldsymbol{\beta}_j \|_2^2 \\
&s.t. \; \frac{1}{n}\boldsymbol{\theta}^T\mathbf{Y}^T\mathbf{Y}\boldsymbol{\theta}=1, \; \boldsymbol{\theta}^T\mathbf{Y}^T\mathbf{Y}\boldsymbol{\theta}_l=0 \quad \forall l=1,2, \dots, j-1.
\end{align*}

\end{enumerate}

For computational implementation details of the above steps we refer the reader to \cite{SLDA} and \cite{spasm}. 

The solution of problem (\ref{SDA}) provides $q$ discriminant direction, 
$\boldsymbol{\beta}_1,\boldsymbol{\beta}_2,...,\boldsymbol{\beta}_q$, allowing standard LDA procedure with the $n \times q$ matrix  $\left(\boldsymbol{X\beta}_1 \; \boldsymbol{X\beta}_2\; \dots \; \boldsymbol{X\beta}_{q}\right)$ to be performed.

Solving Eq. (\ref{enet}) involves the well-known elastic-net problem (e-net) \cite{enet}, which is similar to LASSO (least absolute shrinkage and selection operator) \cite{lasso}, since e-net performs automatic sparse variable selection and it also allows selection of groups of correlated variables (this is due to the introduction of $\ell_2$-norm term). 

In this work we propose the use of a penalized version of the SDA method to efficiently solve the binary classification problem appearing in BCI systems based on ERPs. This new method seeks to increase classification performance by taking into account information about the difference between classes by means of the inclusion of appropriate anisotropy matrices into the penalizing terms. The use of adaptive penalizers and, in particular of anisotropy matrices in regularization method for inverse ill-posed problems is a new approach that has shown to produce significantly better results than those obtained with the corresponding non-adaptive or isotropic penalizers \cite{mazzieri2015mixed}. 

\section{A new approach: Kullback-Leibler Penalized Sparse Discriminant Analysis}
\label{ourKLSDA}
In a pattern recognition problem is very important to analyze the data and be able to extract from them as much prior information as possible since, needless to say, a good classification performance will largely depend on how well the problem is understood  through the available data.   

Let $\mathbf{X}$, $\mathbf{Y}$, $\boldsymbol{\beta}_j$, $\boldsymbol{\theta}_j$, $\lambda_1$ and $\lambda_2$ be as before, and let $\mathbf{D}_1$ and $\mathbf{D}_2$ be $p \times p$ diagonal positive definite matrices. The KLSDA scheme consists of recursively solving, for $j=1, 2, \dots, q$, the following regularized constrained least squares problem:
\begin{align}
\nonumber
\left(\boldsymbol{\beta}_j, \boldsymbol{\theta}_j\right)&=\argmin_{\boldsymbol{\beta}\in \mathbb{R}^p, \boldsymbol{\theta}\in \mathbb{R}^K} \{ \| \mathbf{Y}\boldsymbol{\theta}- \mathbf{X}\boldsymbol{\beta} \|_2^2 + \lambda_1 \|\mathbf{D}_1 \boldsymbol{\beta}\|_1 + \lambda_2\|\mathbf{D}_2\boldsymbol{\beta}\|_2^2 \}, \\
& s.t. \quad \frac{1}{n}\boldsymbol{\theta}^T\mathbf{Y}^T\mathbf{Y}\boldsymbol{\theta}=1, \; \boldsymbol{\theta}^T\mathbf{Y}^T\mathbf{Y}\boldsymbol{\theta}_l=0 \quad \forall l=1,2, \dots, j-1.
\label{KLSDA}
\end{align}
Here again, for $j=1$ the orthogonality condition is vacuous. As in the SDA case, the solution of problem (\ref{KLSDA}) is approximated by alternatively iterating the following two steps (with an adequate initialization):

\begin{enumerate}
\item[1.] Given $\boldsymbol{\theta}_j$ solution of (\ref{lalala}), solve:
\begin{equation}
\boldsymbol{\beta}_j= \argmin_{\boldsymbol{\beta} \in \mathbb{R}^p} \{ \| \mathbf{Y}\boldsymbol{\theta}_j- \mathbf{X}\boldsymbol{\beta} \|_2^2 +\lambda_1\|\mathbf{D}_1\boldsymbol{\beta}\|_1 + \lambda_2 \|\mathbf{D}_2\boldsymbol{\beta}\|_2^2 \}.
\label{Genet}
\end{equation}
\item[2.] Given $\boldsymbol{\beta}_j$ solution of (\ref{Genet}), solve:
\begin{align}
\nonumber
\boldsymbol{\theta}_j&=\argmin_{\boldsymbol{\theta} \in \mathbb{R}^K} \| \mathbf{Y}\boldsymbol{\theta}- \mathbf{X}\boldsymbol{\beta}_j \|_2^2 \\
&s.t. \quad \frac{1}{n}\boldsymbol{\theta}^T\mathbf{Y}^T\mathbf{Y}\boldsymbol{\theta}=1, \; \boldsymbol{\theta}^T\mathbf{Y}^T\mathbf{Y}\boldsymbol{\theta}_l=0 \;\forall l=1,2, \dots, j-1.
\label{lalala}
\end{align}
\end{enumerate}

The vector $\boldsymbol{\beta}_j$ solution of (\ref{Genet}), not only inherits both the correlated variables selection and  sparsity properties of SDA, but it also contains in each one of its components appropriate discriminative information which is suitable for improving separability between classes. As before, the classification rule is constructed based upon the $n \times q$ matrix $\left(\boldsymbol{X\beta}_1 \; \boldsymbol{X\beta}_2\; \dots \; \boldsymbol{X\beta}_{q}\right)$. In the following subsection we show how the Kullback-Leibler divergence can be used for constructing the anisotropy matrices $\mathbf{D}_1$ and $\mathbf{D}_2$ in such a way that they properly incorporate discriminative information into KLSDA.

\subsection{Kullback-Leibler anisotropy matrix}
\label{KLAniso}
Discriminative information can be incorporated into KLSDA by appropriately quantifying the ``distances'' between classes, or more precisely, between their probability distributions. Although there is a wide variety of ``metrics'' for comparing probability distributions \cite{Distances}, we shall use here the well-known Kullback-Leibler divergence \cite{kullback}. The decision to use this particular ``metric'' is due not only to its nice mathematical properties, but also to the fact that it was already successfully applied in many classification problems \cite{automatic, kullbackSVM, KullbackClassification}. 

Let us suppose first that we are dealing with a binary classification problem ($K=2$). Let  $\mathbf{N}$ be a discrete random variable defined on a discrete outcome space $\mathcal{N}$ and consider two probability functions $f_1(n)$ and $f_2(n)$, $n\in \mathcal{N}$. Then, the Kullback-Leibler ``distance'' (KLD) of $f_1$ relative to $f_2$ is defined as:
\[
D_{\text{KL}}(f_1|| f_2)\dot=\sum_{n\in \mathcal{N}} f_1(n) \log \left( \frac{f_1(n)}{f_2(n)} \right),
\]
with the convention that $0.\log 0\dot=0$. Although $D_{\text{KL}}(f_1|| f_2)$ quantifies the discrepancy between $f_1$ and $f_2$, it is not a metric in the rigorous mathematical sense, because it is not symmetric and it does not satisfy the triangle inequality. If, for any reason, symmetry is desired then a modified KLD, called J-divergence, can be defined as follows:
\[
J_{\text{KL}}(f_1, f_2)\dot=\frac{D_{KL}(f_1|| f_2) + D_{KL}(f_2|| f_1)}{2}.
\]

For measuring discrepancy between $K$ probability distributions, $f_{1}, \dots, f_{K}$, since KLD is ``additive'' \cite{kullback}, an appropriate measure can be build up by adding up the J-divergences between all possible pairs of distributions, that is:
\[
J_{KL}\left(\left\lbrace f_j \right\rbrace_{j=1}^K \right)\dot=\sum_{i=1}^{K-1}\sum_{j=i+1}^{K} J_{\text{KL}}(f_i,f_j).
\]

Let $f_j^i(\cdot)$ be the probability function of the $j^{th}$ class in the $i^{th}$ sample, with $j=1,2,\dots,K$ and $i=1,2,\dots, p$. We define the J-divergence at sample $i$ as
\begin{equation}
J_{KL}(i)\dot = J_{KL}\left(\left\lbrace f_j^i\right\rbrace_{j=1}^K \right).
\label{JKL}
\end{equation}
This function quantifies the discrepancy between the $K$ classes at sample $i$. A value of $J_{KL}(i)$ close to zero means that there is very little discriminative information at sample $i$, while a large value of $J_{KL}(i)$ means that sample $i$ contains a significant amount of discriminative information which we definitely want to take into account for constructing the solution vectors $\boldsymbol{\beta}_j$. In Section \ref{resultsKLD} we show in detail how J-divergence is able to highlight the most discriminative samples. 

The available a priori discriminative information can be incorporated into the KLSDA formulation (Eq. (\ref{KLSDA})) by means of appropriately constructed anisotropy matrices $\mathbf{D}_1$ and $\mathbf{D}_2$. Since we wish to stand out those samples containing significant amounts of discriminative information, the matrices $\mathbf{D}_1$ and $\mathbf{D}_2$ must be constructed so as to strongly penalize those samples where there is little or none discriminative information while avoiding penalization at the remaining ones. 

\subsection{Computational implementation}
Our computational implementation of KLSDA bellow is made by appropriately modifying the original 
SDA algorithm \cite{spasm}. Thus KLSDA is mainly solved in two steps. In the first step, Eq. (\ref{Genet}), which is a generalized version of the e-net problem \cite{Genet}, is solved. The second step consists of updating the optimal score vector $\boldsymbol{\theta}_j$ by solving Eq. (\ref{lalala}). It is shown in \cite{SLDA} that the solution of Eq. (\ref{lalala}) is given by $\boldsymbol{\theta}_j=s(\mathbf{I}-\mathbf{\Theta}_{j-1}\mathbf{\Theta}_{j-1}^T
\boldsymbol{\pi})\boldsymbol{\pi}^{-1}\mathbf{Y}^T\mathbf{X}\boldsymbol{\beta}_j$, where $\boldsymbol{\Theta}$ is the $K \times q$ matrix containing the score vectors $\boldsymbol{\theta}_j$,  $\boldsymbol{\pi}=\frac{1}{n}\mathbf{Y}^T\mathbf{Y}$ and $s$ is a proportionality constant such that $\boldsymbol{\theta}_j\boldsymbol{\pi}\boldsymbol{\theta}_j=1$.

In regard to the first step, it is known that the e-net problem can be reformulated by means of LASSO. In fact by defining the following augmented variables: 
\[\tilde{\mathbf{X}}=\left(
\begin{array}{c}
\mathbf{X}\\
\sqrt{\lambda_2\mathbf{D}_2}\\
\end{array}
\right)_{(n+p)\times p},
\quad
\tilde{\mathbf{Y}}_j=\left(
\begin{array}{c}
\mathbf{Y}\boldsymbol{\theta}_j\\
\mathbf{0}_{p \times 1}\\
\end{array}
\right)_{(n+p)\times 1},
\label{ec1}
\]
the generalized e-net problem (Eq. (\ref{Genet})) can be re-written as:
\begin{equation}
\hat{\boldsymbol{\beta}_j}=\argmin_{\boldsymbol{\beta} \in \mathbb{R}^p} \{ \|\tilde{\mathbf{Y}}_j- \tilde{\mathbf{X}}\boldsymbol{\beta} \|_2^2 +\lambda_1\|\mathbf{D}_1\boldsymbol{\beta}\|_1  \},
\label{glasso}
\end{equation}
which is known as generalized LASSO \cite{Glasso}. If $\mathbf{D}_1$ is invertible the solution of (\ref{glasso}) can be found as $\hat{\boldsymbol{\beta}_j}=\mathbf{D}_1^{-1}\hat{\boldsymbol{\alpha}}_j$, where $\hat{\boldsymbol{\alpha}}_j$ is the solution of:
\begin{equation}
\hat{\boldsymbol{\alpha}_j}=\argmin_{\boldsymbol{\alpha} \in \mathbb{R}^p} \{ \|\tilde{\mathbf{Y}}_j- \tilde{\mathbf{X}}\mathbf{D}_1^{-1}\boldsymbol{\alpha} \|_2^2 +\lambda_1\|\boldsymbol{\alpha}\|_1  \}.
\label{ec2}
\end{equation}

The LARS-EN algorithm, presented in \cite{enet}, provides an efficient way of solving problem (\ref{ec2}).
\subsection{Regularization parameters}
It is well-known that in every regularization method the choice of the regularization parameters is crucial. For Tikhonov-type functionals a popular and widely used method for approximating the optimal parameters is the so called L-curve criterion. One of the main advantages of this selection criterion is the fact that it does not require any prior knowledge about the noise. Roughly speaking, the method finds an optimal compromise between the norm of the residual and the norm of the regularized solution by selecting the point of maximal curvature in a log-log plot of those two quantities, parametrized by the regularization parameter. For details see \cite{lcurve}. 

Despite its popularity, the L-curve method cannot be directly applied to multi-parameter penalization functionals like (\ref{Genet}). In 1998, Belge et al. proposed and extension of the L-curve technique for approximating the optimal regularization parameters in those cases, called L-hypersurface technique \cite{hyperlcurve}, which we briefly describe next.

Let $\mathbf{X} \in \mathbb{R}^{n \times p}$, $\mathbf{y} \in \mathbb{R}^n$, $\mathbf{R}_i \in \mathbb{R}^{l \times p}$ for $i=1, \dots, M$, be all given, and consider the following multi-parameter regularized least squares problem:
\begin{equation}
\boldsymbol{\beta}(\boldsymbol{\lambda})=\text{arg} \min_{\boldsymbol{\beta}}\{\|\mathbf{y}-\mathbf{X}\boldsymbol{\beta} \|_2^2 + \sum_{i=1}^M \lambda_i\|\mathbf{R}_i\boldsymbol{\beta}\|_r^r \},
\label{Lhyper}
\end{equation}
where $1\leq r \leq 2$ and let $\boldsymbol{\lambda}=(\lambda_1, \dots, \lambda_M)$ denote the regularization parameter vector. Define $z(\boldsymbol{\lambda})\dot =\log \|\mathbf{y}-\mathbf{X}\boldsymbol{\beta}(\boldsymbol{\lambda)}\|_2^2$ and $x_i(\boldsymbol{\lambda})\dot =\log \|\mathbf{R}_j\boldsymbol{\beta}(\boldsymbol{\lambda)}\|_r^r$, for $i=1,\dots,M$. The L-hypersurface associated to problem (\ref{Lhyper}) is the subset $S(\boldsymbol{\lambda})$ of $\mathbb{R}^{M+1}$ parametrized by $\boldsymbol{\lambda}\in \mathbb{R}_+^M$ defined as 
$S(\boldsymbol{\lambda}) \dot=\{(x_1(\boldsymbol{\lambda}),..., x_M(\boldsymbol{\lambda}),$ $z(\boldsymbol{\lambda} )): \boldsymbol{\lambda} \in \mathbb{R}_+^M \}$.

The L-hypersurface criterion consists of finding $\boldsymbol{\lambda}^* \in \mathbb{R}_+^M$ such that $S(\boldsymbol{\lambda^*})$ is the point of maximal Gaussian curvature of the L-hypersurface $S(\boldsymbol{\lambda})$. Although approximating $\boldsymbol{\lambda}^*$ is most of the times very costly from a computational point of view, in \cite{hyperlcurve} the authors show that a good approximation to $\boldsymbol{\lambda}^*$ is given by the minimizer of the residual norm, i.e., by the vector $\hat{\boldsymbol{\lambda}} \in \mathbb{R}_+^M$ satisfying $z(\hat{\boldsymbol{\lambda}})=\min_{\boldsymbol{\lambda \in \mathbb{R}_+^M}}z(\boldsymbol{\lambda})$.
  
We describe next the details of how the optimal regularization vector $(\hat{\lambda}_1, \hat{\lambda}_2)$ given by the L-hypersurface approach is approximated within the KLSDA context  formalized in  Eq. (\ref{KLSDA}) or, more precisely, in the context of the generalized e-net problem (Eq. (\ref{Genet})). For $j=1, 2, \dots, q$, define $z_j(\boldsymbol{\lambda})=\log \|\mathbf{Y}\boldsymbol{\theta}_j-\mathbf{X}\boldsymbol{\beta}_j(\boldsymbol{\lambda})\|_2^2$, $x_{1,j}(\boldsymbol{\lambda})=\log \|\mathbf{D}_1\boldsymbol{\beta}_j(\boldsymbol{\lambda})\|_1$ and $x_{2,j}(\boldsymbol{\lambda})=\log \|\mathbf{D}_2\boldsymbol{\beta}_j(\boldsymbol{\lambda})\|_2^2$. The L-hypersurface associated to (\ref{KLSDA}) is then defined as $S_j(\boldsymbol{\lambda})= \{( x_{1,j}(\boldsymbol{\lambda} ),x_{2,j}(\boldsymbol{\lambda}),$ $z_j(\boldsymbol{\lambda})) : \boldsymbol{\lambda}\in \mathbb{R}_+^2\}$. 

Although generalized e-net is defined in terms of $\lambda_1$ and $\lambda_2$, there are other possible choices for tuning parameters \cite{enet}. For example, the $\ell_1$-norm of the coefficients ($t$) can be chosen instead of $\lambda_1$, In fact, this can be achieved by re-writing the LASSO version (Eq. (\ref{ec2})) of our generalized e-net as a constrained optimization problem with an upper bound on $\|\boldsymbol{\alpha}\|_1$. Similarly, since the LARS-EN algorithm is a forward stagewise additive fitting procedure, the number of steps $\kappa$ of the algorithm can also be used as a tuning parameter replacing $\lambda_1$. This is so because, for each fixed $\lambda_2$, LARS-EN produces a finite number of vectors $\boldsymbol{\beta} \in \mathbb{R}^p$ which are approximations of the true solution at each step $\kappa$. In our numerical experiment we adopted $\lambda_2$ and $\kappa$ as tuning parameters and, in accordance to Belge's remark described above, the best $\boldsymbol{\beta}_j$ solutions were selected as those minimizing the residual norm. All the steps of the algorithm solving KLSDA with this proposal (together with automatic parameter selection) are presented in Algorithm \ref{KLSDAalgSup}. 

\begin{algorithm}
\caption{KLSDA with automatic parameter selection}
\label{KLSDAalgSup}
\begin{algorithmic}[1]
\INPUT $\mathbf{X}$, $K$, $q$, $\mathbf{Y}$ , $\mathbf{D}_1$, $\mathbf{D}_2$, $\Lambda_2=\left\lbrace \lambda_2^{(1)}, \dots, \lambda_2^{(d)}  \right\rbrace$.
\State Define $\boldsymbol{\pi}=\frac{1}{n}\mathbf{Y}^T\mathbf{Y}$
\State Initialize: $\mathbf{\Theta}=eye(K,q)$.
%\State Let $\mathbf{B}_j$ be the solution path, $\boldsymbol{\beta}^{(1)}_j, \dots, \boldsymbol{\beta}^{(\kappa)}_j$, for a given $\lambda_2$ 
\For {$j=1, \dots, q$}
\While {sparse discriminative direction $\boldsymbol{\beta}_j$ has not converged} 
\For {$i=1, \dots, d$}
\State Re-define the variables: 
\begin{equation*}
\tilde{\mathbf{X}}=\left(
\begin{array}{c}
\mathbf{X}\\
\sqrt{\lambda_2^{(i)}\mathbf{D}_2}\\
\end{array}
\right)_{(n+p)\times p},
\quad
\tilde{\mathbf{Y}}_j=\left(
\begin{array}{c}
\mathbf{Y}\boldsymbol{\theta}_j\\
\mathbf{0}_{p \times 1}\\
\end{array}
\right)_{(n+p)\times 1}
\end{equation*}
\State Solve the generalized e-net problem and save the solution path:
\begin{equation*}
(\mathbf{A}_j, \kappa)=\text{LARSEN}(\tilde{\mathbf{X}}_j\mathbf{D}_1^{-1}, \tilde{\mathbf{Y}}_j), \quad \mathbf{B}_j=\mathbf{D}_1^{-1}\mathbf{A}_j
\end{equation*}
\State Find the residual: 
\begin{center}
$\mathbf{R}(\lambda_2, 1:\kappa)=\|\mathbf{Y}\boldsymbol{\theta}_j-\mathbf{X}\mathbf{B}_j\|_2^2$
\end{center}
\State Save the solutions:
\begin{center}
$\mathbf{B}_{\text{all}}(\lambda_2, 1:\kappa, :)=\mathbf{B}_j$
\end{center}
\EndFor
\State Select the optimal direction:
\begin{align*}
(\hat{\lambda_2}, \hat{\kappa})&= \argmin_{\lambda_2, \kappa} \mathbf{R}(\lambda_2, \kappa)\\
\boldsymbol{\beta}_j&= \mathbf{B}_{\text{all}}(\hat{\lambda_2},\hat{\kappa},:)
\end{align*}

\State Update $j^{th}$ column of $\mathbf{\Theta}$:
$\boldsymbol{\theta}_j=(\mathbf{I}-\mathbf{\Theta}_{(j-1)}\mathbf{\Theta}_{(j-1)}^T
\boldsymbol{\pi})\boldsymbol{\pi}^{-1}\mathbf{Y}^T\mathbf{X}\boldsymbol{\beta}_j$
\State Normalize to unit length: $\boldsymbol{\theta}_j=\frac{\boldsymbol{\theta}_j}{\sqrt{\boldsymbol{\theta}_j^T\boldsymbol{\pi}\boldsymbol{\theta}_j}}$
\EndWhile
\EndFor
\OUTPUT: $\left[\boldsymbol{\theta}_1, \dots, \boldsymbol{\theta}_q\right]$, $\left[\boldsymbol{\beta}_1, \dots, \boldsymbol{\beta}_q\right]$
\end{algorithmic}
\end{algorithm}

\section{P300 speller databases}
\label{datasets}
Two real ERP-EEG databases were used to evaluate the classification performance of our KLSDA method. 
\subsection{Dataset-1}
Dataset-1 is an open-access P300 speller database from Laboratorio de Investigación en Neuroimagenología at Universidad Autónoma Metropolitana, Mexico D.F., 
described in \cite{base-datos}. This database consists of EEG records acquired from 25 healthy subjects, recorded by 10 channels (Fz, C3, Cz, C4, P3, Pz, P4, PO7, PO8, Oz) at 256 Hz sampling rate using the g.tec gUSBamp. During the acquisition, the EEG records were filtered with a Chebyshev Notch 4$^{th}$ order filter with cutoff frequencies of 58-62 Hz and a Chebyshev band-pass 8$^{th}$ order filter with cutoff frequencies of 0.1-60 Hz. A 6-by-6  matrix containing letters and numbers was presented to each subject on a computer screen. During the experiment, the subject was asked to spell different words. The person had to focus on one character at the time. As stimulus, a row or a column of the matrix was randomly highlighted for a period of 62.5 ms with inter-stimuli intervals of 125 ms. In each stimulating block (consisting in 12 consecutive flashings), every row and every column of the matrix was intensified only once. For each character to be spelled this stimulating block was repeated 15 times. If the person was well concentrated, a relevant event occurred when the chosen character was illuminated, i.e. an ERP was elicited. Thus, in a binary classification problem, the 6-by-6 matrix generates twelve possible events, of which only two are labelled as target. 

Each subject participated in 4 sessions, of which the first two were copy-spelling runs, i.e, they contained the true label data vector, with 21 characters to be spelled. For this reason, in the present work we used those two copy-spelling sessions as our dataset. The EEG records were filtered from 0.1 Hz to 12 Hz by a 4$^{th}$ order forward-backward Butterworth band-pass filter. A 1000 ms data segment was extracted (windowed) from the EEG records at the beginning of each stimulus. A total of 3780 EEG patterns (of which only 630 are target) with dimensionality of $10 \times 256=2560$ conforms the dataset per each subject.
\subsection{Dataset-2}
Dataset-2 is a P300 speller database of patients with amyotrophic lateral sclerosis (ALS) obtained from the Neuroelectrical Imaging and BCI Laboratory, IRCCS Fondazione Santa Lucia, Rome, Italy. The EEG data were recorded using g.MOBILAB and g.Ladybird with active electrode (g.tec, Austria) from eight channels (Fz, Cz, Pz, Oz, P3, P4, PO7 and PO8). All channels were referenced to the right earlobe and grounded to the left mastoid. The EEG signals were digitized at 256 Hz. Eight participants with ALS were required to copy-spell seven predefined words of five characters each using a P300 speller paradigm. As in Dataset-1, a 6-by-6 matrix containing alphanumeric characters was used. The rows and columns of the matrix were randomly intensified for 125 ms, follow by an inter stimulus interval of the same length. In each stimulating block all rows and columns were intensified 10 times. For more details about this dataset we refer the reader to \cite{P300ALS}. 

In a pre-processing stage, the EEG records from each channel were band-pass filtered from 0.1 Hz to 10 Hz using a 4$^{th}$ order Butterworth filter. Then, data segments of 1000 ms were extracted at the beginning of each intensification. The ALS P300 speller dataset consists of 4200 patterns, of which 700 are target, with $256 \times 8=2048$ sample points. 

\section{Experiments and results}
\label{Experiments}
In this section we show how the KLSDA method is implemented in the context of the aforementioned real ERP-based BCI classification problem with both datasets described in Section \ref{datasets}. Since we are dealing here with a binary classification problem ($K=2$), there is only one direction vector $\boldsymbol{\beta}$ in Eq. (\ref{KLSDA}), with which a linear classifier has then to be implemented with the $n \times 1$ projected data vector $\mathbf{X}\boldsymbol{\beta}$. 

In the computational implementation of Algorithm \ref{KLSDAalgSup}, the parameter $\lambda_2$ was allowed to vary between $10^{-8}$ and $10^{-1}$ in a log-scale, and an upper bound of 800 was chosen for the $\ell_1$-norm of the coefficients ($t\leq 800$). All codes were run in MATLAB on an AMD FX(tm)-6300 Six-Core PC with 4GB of memory. It is timely to mention here that although the CPU-time needed to find the direction vector was around 2800 secs, once $\boldsymbol{\beta}$ was found, training and testing the linear classifier with the projected data vector $\mathbf{X}\boldsymbol{\beta}$ took only about 3.36 secs.  

In experiments we decided to use the symmetric version of KLD, $J_{KL}$, as a measure of discrepancy. The probability distribution of each class was estimated by using histogram with centered bins between the most minimum and most maximum value in target and non-target training data. We constructed an anisotropy matrix $\mathbf{D}=diag(d_1, \dots, d_p)$, where $d_i=C/J_{KL}(i), \; i=1, \dots, p$, with $C$ being the constant that makes $\text{det}(\mathbf{D})=1$, i.e., $C=\left(\prod_{j=1}^p J_{KL}(j)\right)^{1/p}$. Note that with this choice, $d_i$ is large where $J_{KL}(i)$ is small and vice-versa. If there exist $i_0, 1\leq i_0 \leq p$, such that $J_{KL}(i_0)=0$ then the matrix $\mathbf{D}$ cannot be formally defined as above. This case, however, can be overcomed by simply replacing $J_{KL}(i_0)$ by $J_{KL}(i_0)+\epsilon$, for $i=1, \dots, p$, with $\epsilon$ very small. The anisotropy matrix $\mathbf{D}$ so defined is clearly diagonal, symmetric and positive definite. 

Four different configurations of KLSDA, denoted by KLSDA0, KLSDA1, KLSDA2 and KLSDA3, were implemented. The first one KLSDA0 correspond to the standard SDA approach with automatics parameter selection.  The second and third ones, KLSDA1 and KLSDA2, incorporate the anisotropic matrix $\mathbf{D}$ in the $\ell_1$ and $\ell_2$ penalizers, respectively. Finally, KLSDA3 incorporate the anisotropic matrix $\mathbf{D}$ in both the $\ell_1$ and $\ell_2$ terms. Table \ref{cuadro} summarizes these four configurations (where $I$ denotes the $p \times p$ identity matrix). The classification performances of KLSDA method is compared with those obtained with the original SDA approach.

\begin{table}[h!]
\caption{KLSDA configurations used.}
\begin{center}
\begin{tabular}{ c  c  c  c  c }
\hline
 & KLSDA0 & KLSDA1 & KLSDA2 & KLSDA3 \\
\hline
 $\mathbf{D_1}$ & $\mathbf{I}$ & $\mathbf{D}$ & $\mathbf{I}$ & $\mathbf{D}$\\
 $\mathbf{D_2}$ & $\mathbf{I}$ & $\mathbf{I}$ & $\mathbf{D}$ & $\mathbf{D}$\\
\hline
\end{tabular}
\end{center}
\label{cuadro}
\end{table}

\subsection{Kullback-Leibler divergence for ERP detection}
\label{resultsKLD}
\begin{figure}
    \centerline{
    \begin{subfigure}[h]{0.48\textwidth}
        \begin{tabular}{c c }
\includegraphics[width=0.45\textwidth]{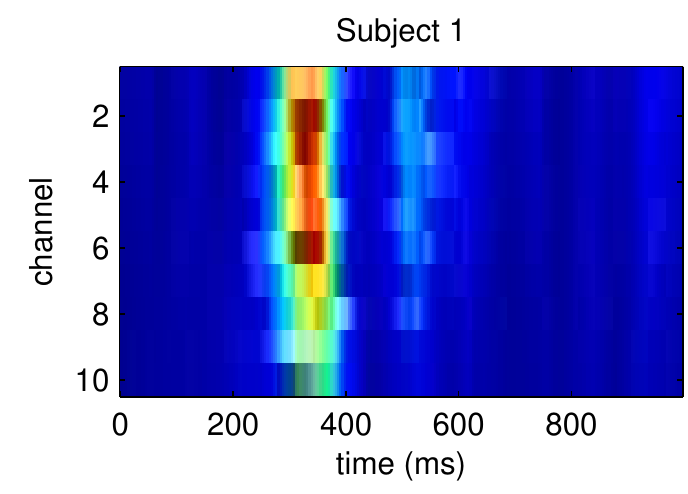} & \includegraphics[width=0.45\textwidth]{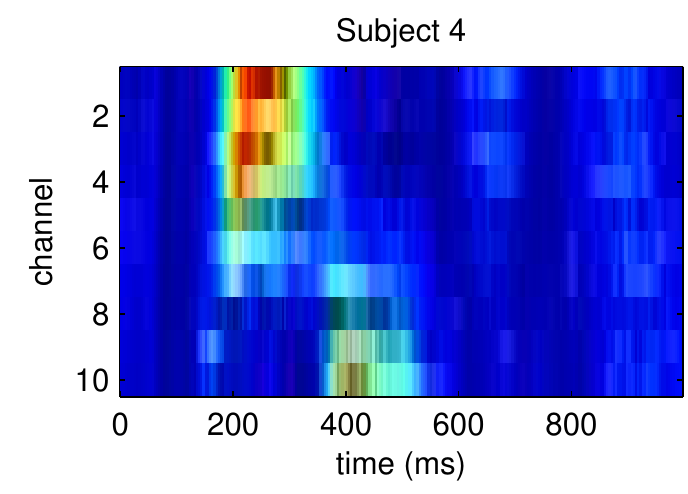}\\
\includegraphics[width=0.45\textwidth]{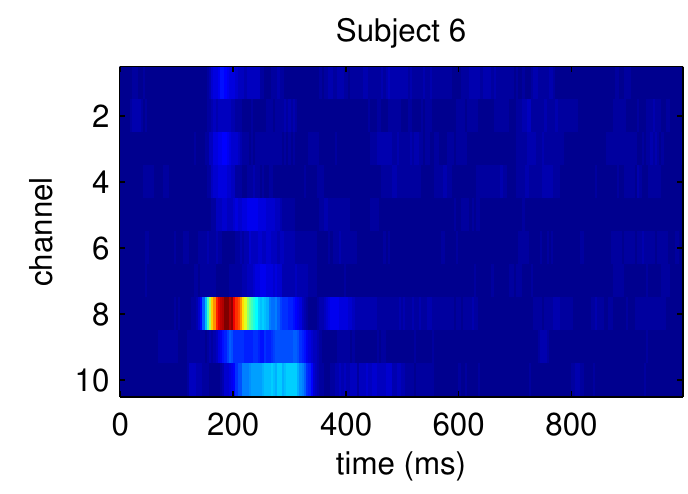} & \includegraphics[width=0.45\textwidth]{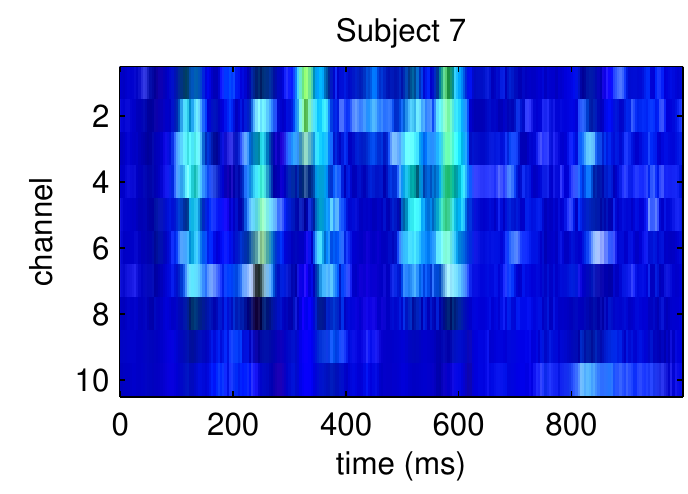}\\
\includegraphics[width=0.45\textwidth]{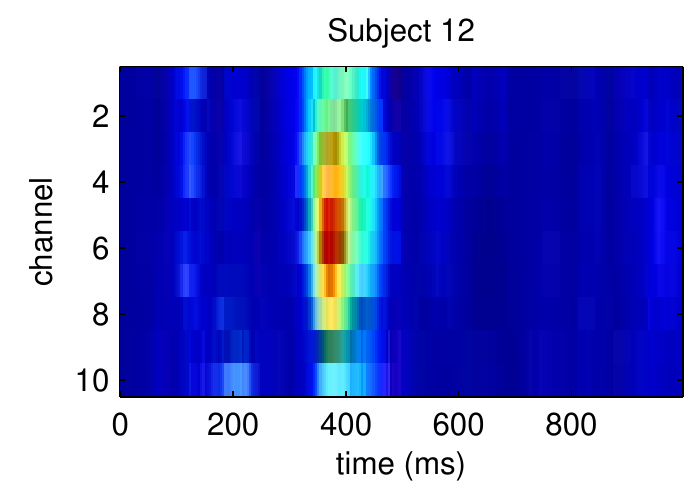} & \includegraphics[width=0.45\textwidth]{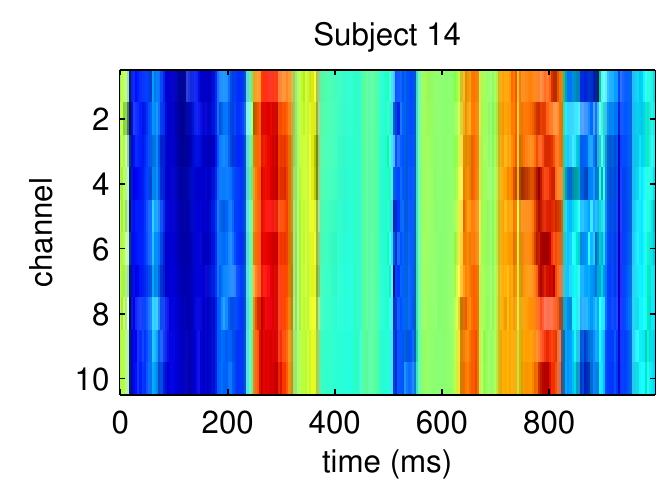}
\end{tabular}
\vspace{0.2in}
        \caption{Dataset-1}
    \end{subfigure}
    \quad
     \begin{subfigure}[h]{0.48\textwidth}
         \begin{tabular}{c c }
\includegraphics[width=0.45\textwidth]{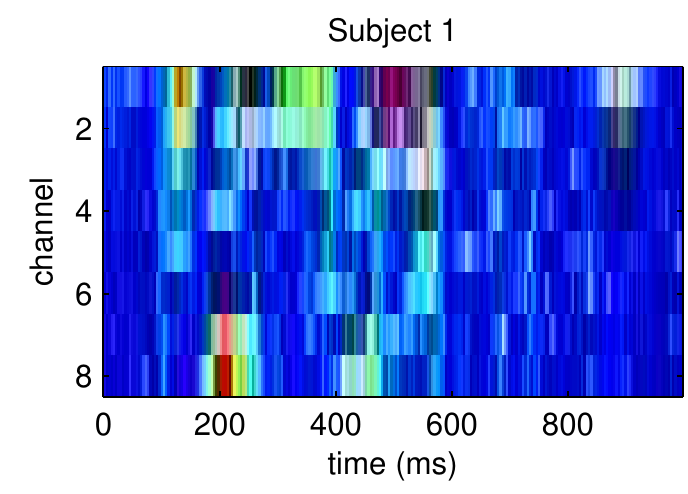} & \includegraphics[width=0.45\textwidth]{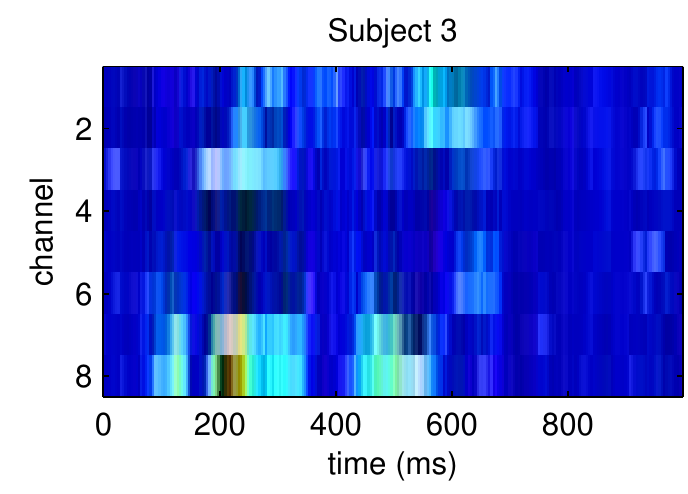}\\
\includegraphics[width=0.45\textwidth]{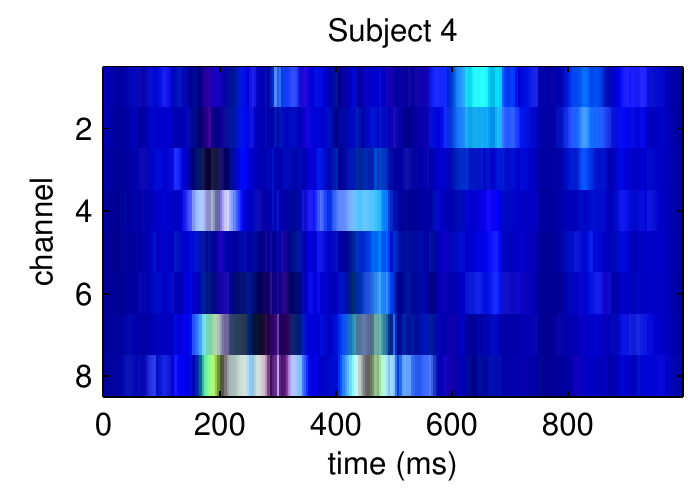} & \includegraphics[width=0.45\textwidth]{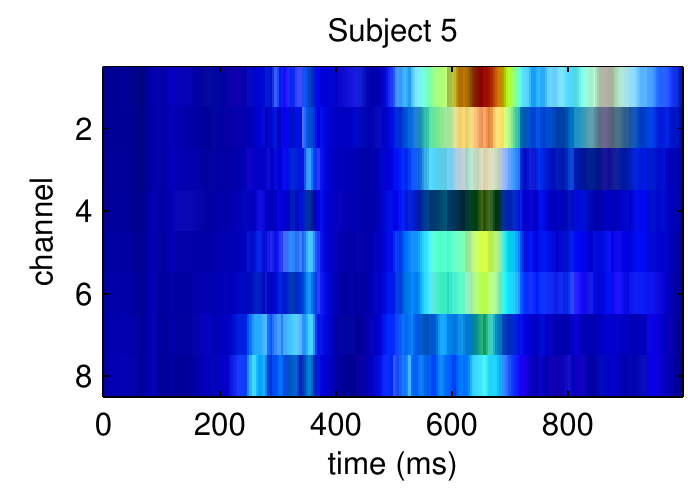}\\
\includegraphics[width=0.45\textwidth]{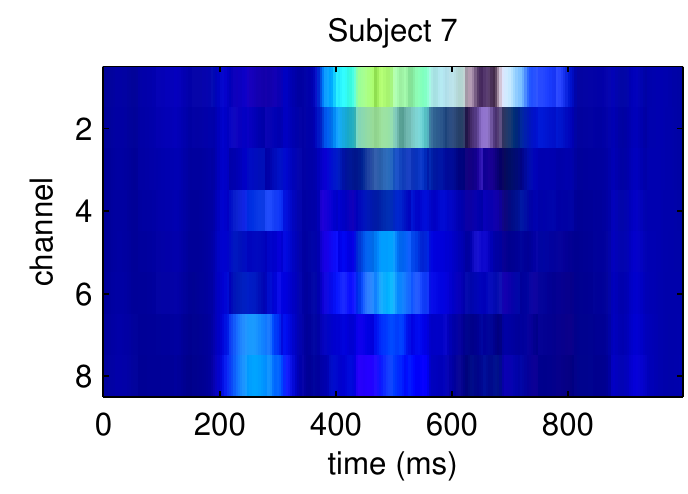} & \includegraphics[width=0.45\textwidth]{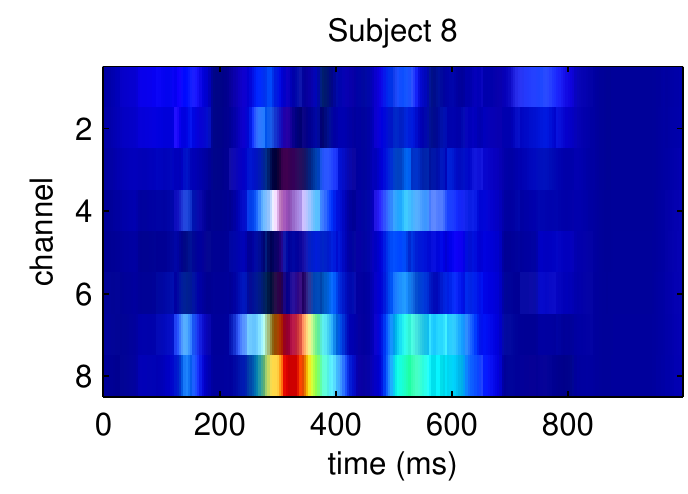}
\end{tabular}
\vspace{0.2in}
        \caption{Dataset-2}
    \end{subfigure}}
     \caption{Discriminative information by means of J-divergence at each sample point in the channel-time plane for six selected subjects from Dataset-1 and Dataset-2.}
     \label{KLD}
\end{figure}

It is reasonable to think that the KLD is an appropriate measure for enhancing the impact of the P300 wave in the KLSDA solution, by selecting both the most discriminative channels and the most discriminative time samples. With this in mind, for both databases, the J-divergence at each sample was estimated as described in Section \ref{KLAniso} (Eq. (\ref{JKL})). An analysis of this J-divergence as a function of channel and time allowed us to detect which samples were the most discriminative ones. The corresponding plots for six selected subjects of Dataset-1 and six selected subjects of Dataset-2 are presented in Figure \ref{KLD}.

Note that although for most of the subjects the discriminative samples are mostly located in the 250-500 ms window (which in fact corresponds to the latency window of the P300 wave), in other cases (e.g. subject 7 and 14 from Dataset-1 and subject 1 from Dataset-2) the most discriminative samples are somewhat randomly distributed all over the plot. Moreover, there are cases in which for some channels the J-divergence shows no contribution at all to class separation (e.g. subject 8 from Dataset-1 and subject 2 from Dataset-2). Figure \ref{KLD} also shows the high variability of the ERP morphology between subjects as it was pointed out in Section \ref{Intro}. 
  
\subsection{Classification results}
There are different available measures for evaluating a BCI classification method \cite{significant}. The receiver operator characteristics (ROC) curve is a powerful tool for evaluating a two-class unbalanced problem \cite{ROC}. In the present work the Area Under the ROC Curve \cite{AUC}, denoted by AUC, was used as the classification performance measure. For avoiding classification bias, a 3-fold cross-validation procedure was implemented. 

\begin{figure}
\centerline{
\includegraphics[width=20cm]{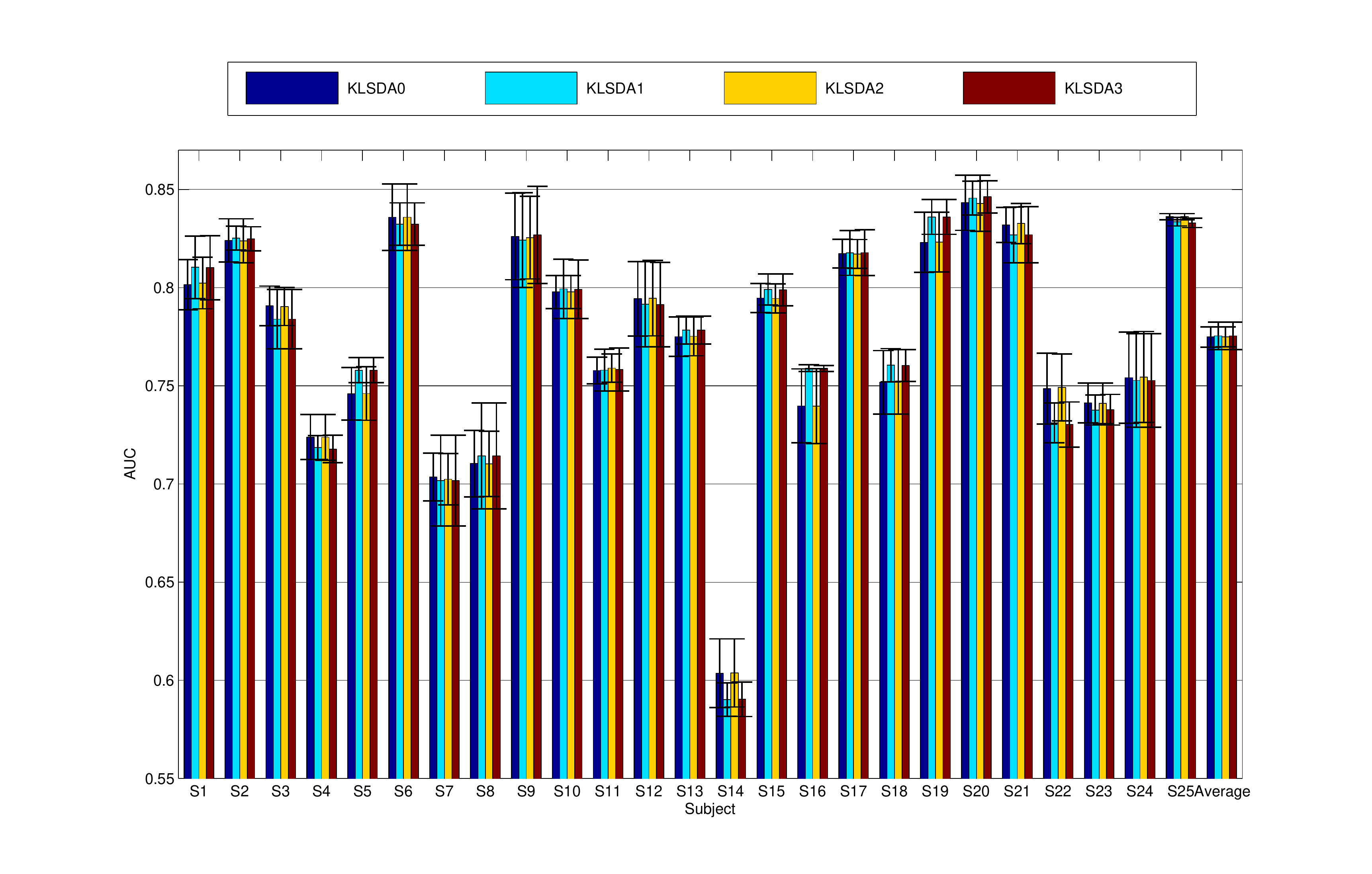}
}
\caption{Area under the 	ROC curve (AUC) on test data from Dataset-1, derived by KLSDA0, KLSDA1, KLSDA2, KLSDA3 and evaluated by 3-fold cross-validation. The errorbar of each subject denotes the standard deviation of AUC on the 3-fold. The errorbar of the average denotes the standard deviation on all subject.} 
\label{clasi}
\end{figure}

Figure \ref{clasi} shows the classification results obtained for each subject from Dataset-1 with the KLSDA0, KLSDA1, KLSDA2 and KLSDA3 methods. Several remarks are in order. First of all, the method which results in the best classification performance seems to be subject dependent. A second observation is that the poorest classification performance corresponds precisely to those subjects with ``randomly spread J-divergence''. A final observation is that in nineteen of the twenty five cases there is at least one  no trivial KLSDA configuration that outperforms the pure SDA (KLSDA0), while in the remaining six cases the performances are essentially equal. 

\begin{figure}
\centerline{
\includegraphics[width=20cm]{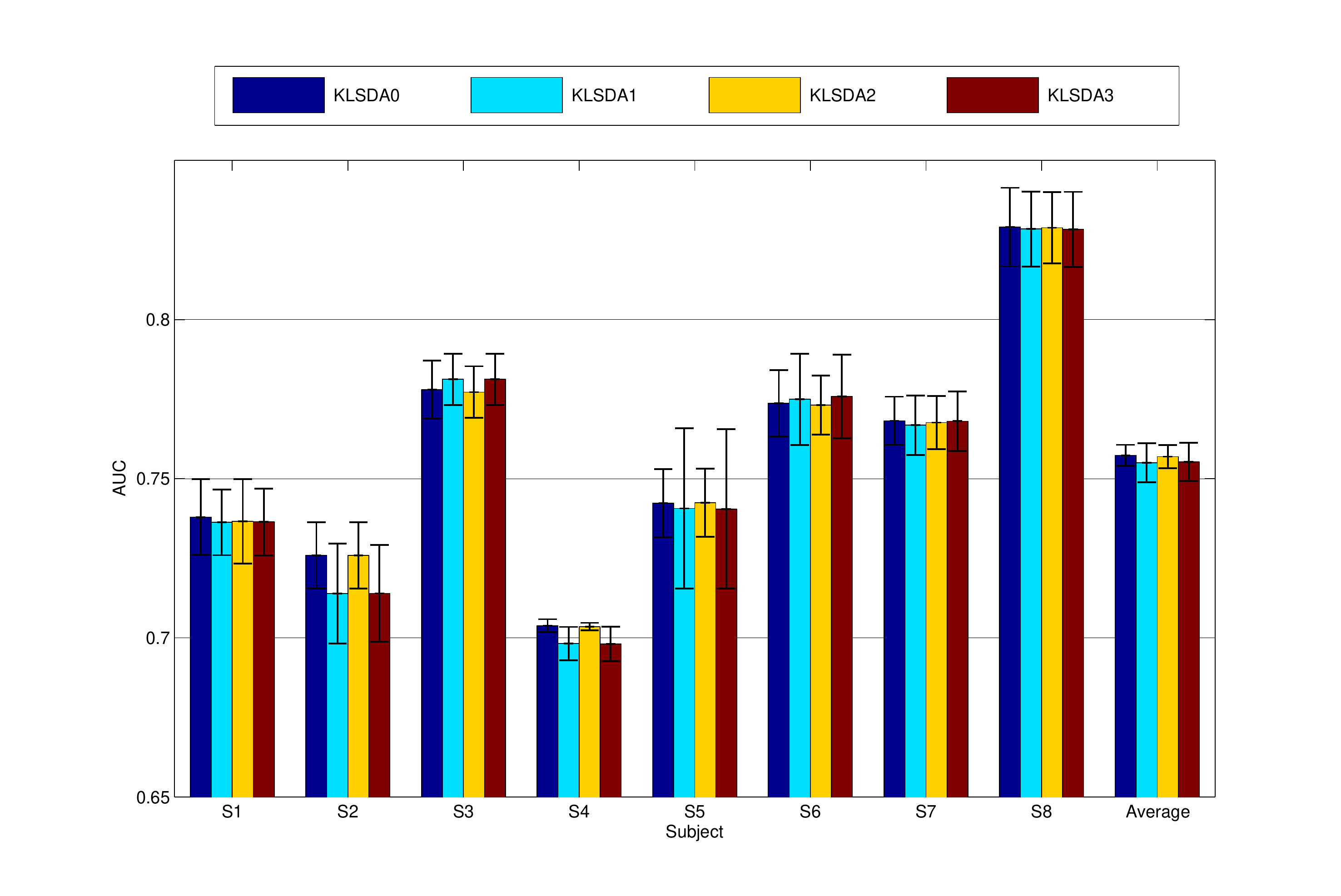}}
\caption{Area under the 	ROC curve (AUC) on test data from each subject of Dataset-2 in each KLSDA configuration. The errorbar of each subject denotes the standard deviation of AUC on the 3-fold. The errorbar of the average denotes the standard deviation on all subject.}
\label{clasi2}
\end{figure}

A similar analysis can be made on the results obtained with Dataset-2 as depicted in Figure \ref{clasi2}. Note that in this dataset, in only three cases one KLSDA configuration outperforms SDA. 

The average classification results for each KLSDA configuration are presented in the last column of Figure \ref{clasi} and Figure \ref{clasi2} for Dataset-1 and Dataset-2, respectively. It is worth noting that for Dataset-1 the average classification result were over 77\% and for Dataset-2 were over 75\% for all KLSDA configuration. These results are very encouraging since an efficient BCI system requires of an accuracy above $70\%$ to allow communication and device control \cite{AdvanceChallenges}. 

We have also used both datasets to test the FLDA classifier with 3-fold cross-validation. As expected, due to the curse-of-dimensionality, the classification results were very poor. In fact, for Dataset-1 the average classification performance was only around 60\% while for Dataset-2 it was near 65\%. These results clearly indicate that regularization improves classification performance.
\subsection{Sparsity analysis}
Given that solution sparsity was desired, the mean of the number of non-zeros values for each KLSDA configuration was analysed. For the KLSDA0, KLSDA1, KLSDA2 and KLSDA3 methods, the mean of the percentage of non-zero values respect to the number of sample points for Dataset-1  were found to be 5.75\%, 7.48\%, 5.78\% and 7.79\%, respectively, while for Dataset-2 those values were found to be 12.27\%, 13.03\%, 12.98\% and 13.19\%. Note that for Dataset-2 sparsity is consistently and significantly lower. This fact most probably reflects the fact that ALS patients database (non-healthy subjects) involved more complex patterns. Finally, it is highly remarkable that all KLSDA configurations with such low percentages of non-zero values achieved such satisfactory classification performances. This observation allows us to conclude that KLSDA constitutes a very robust classifier method. 

\begin{figure}
\begin{tabular}{c c }
\includegraphics[width=0.45\textwidth]{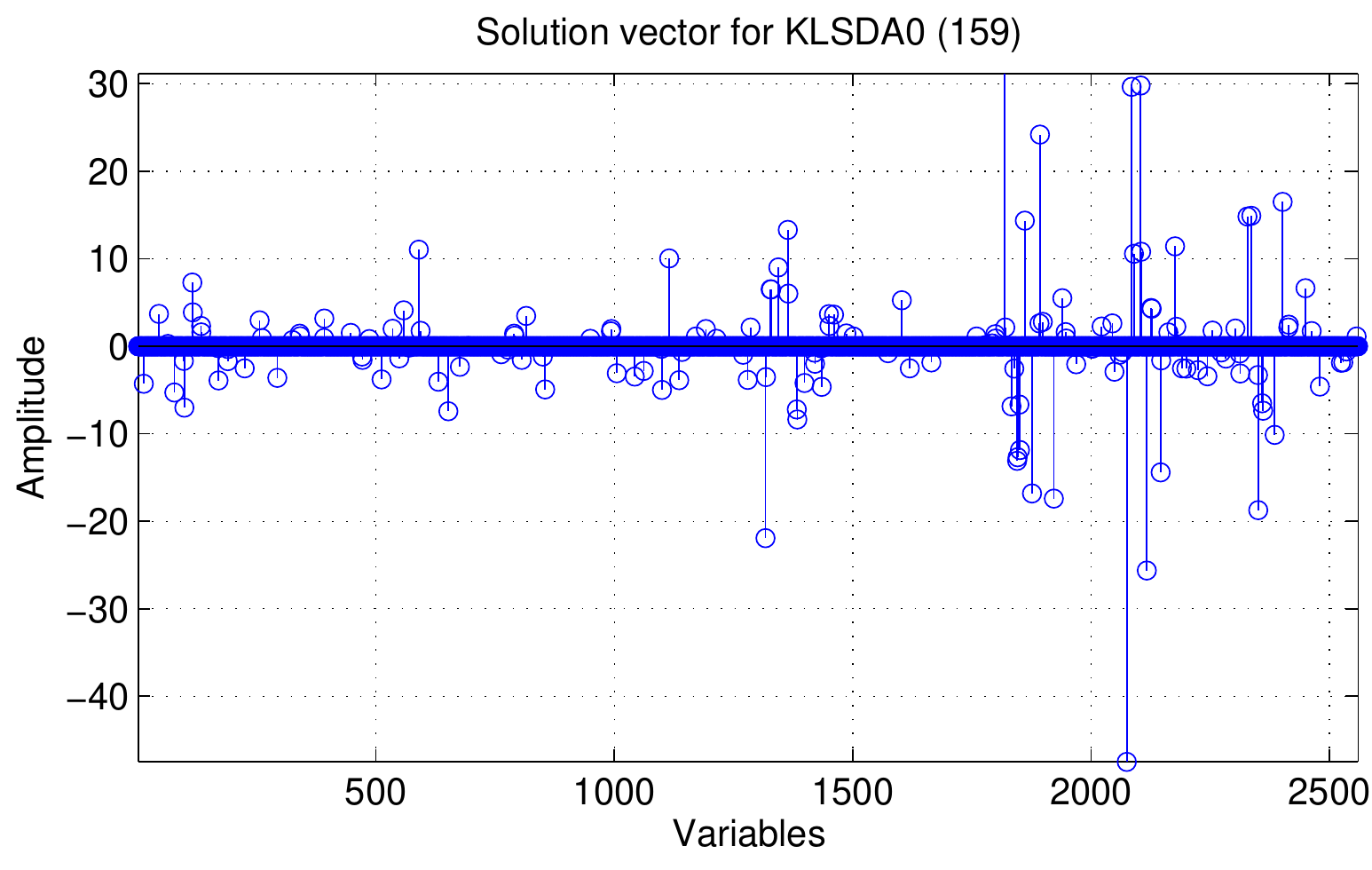} & \includegraphics[width=0.45\textwidth]{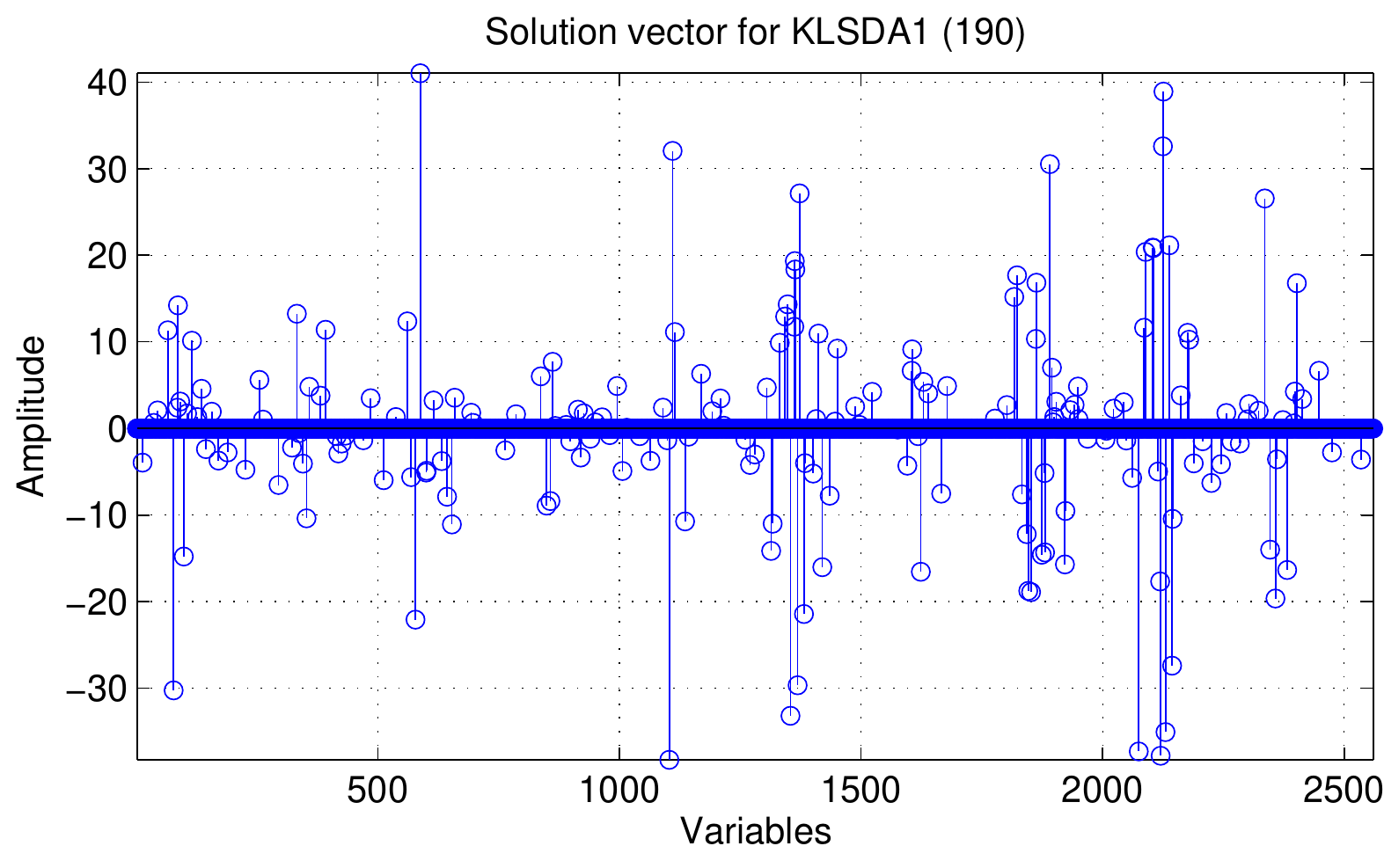}\\
\includegraphics[width=0.45\textwidth]{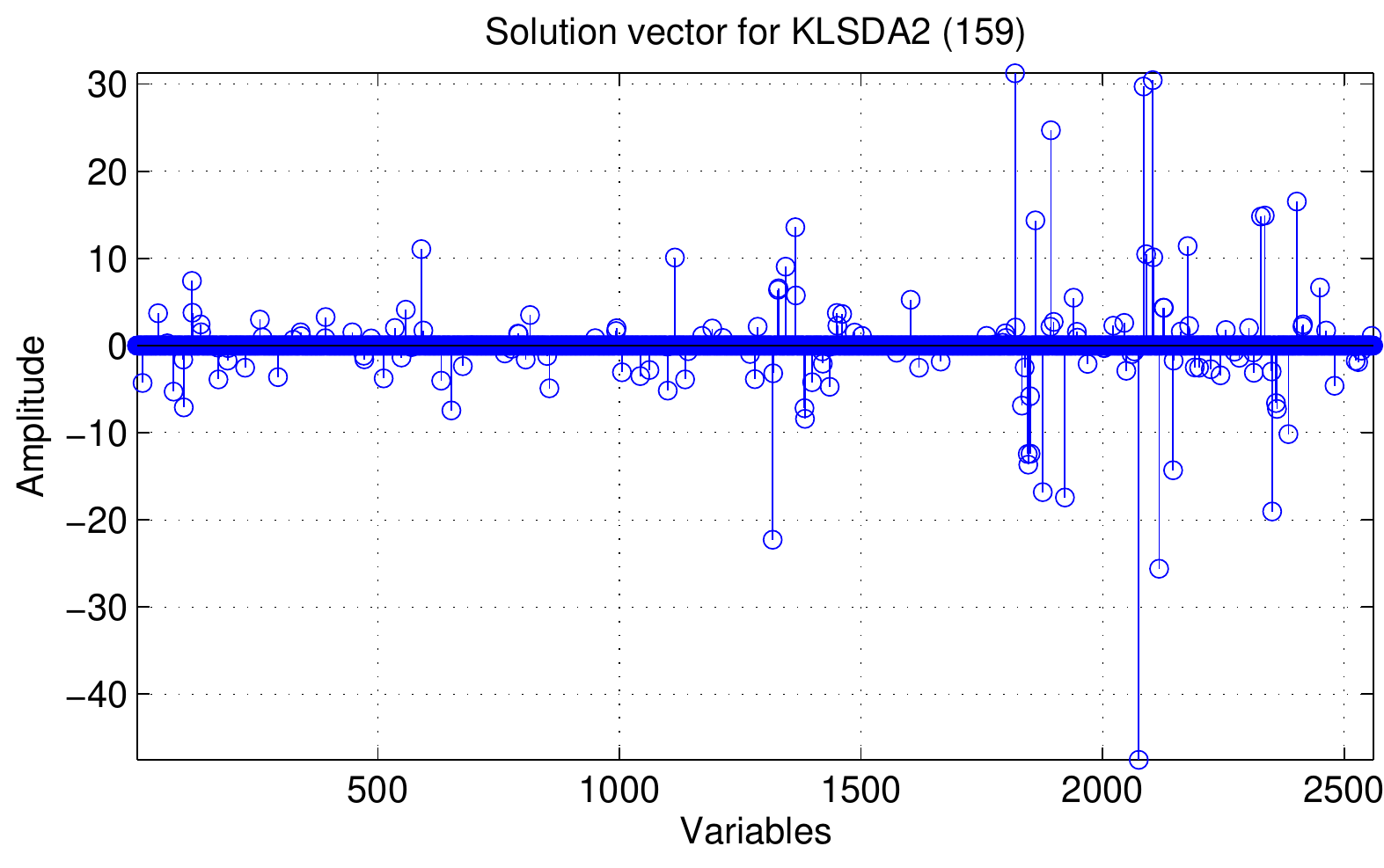} & \includegraphics[width=0.45\textwidth]{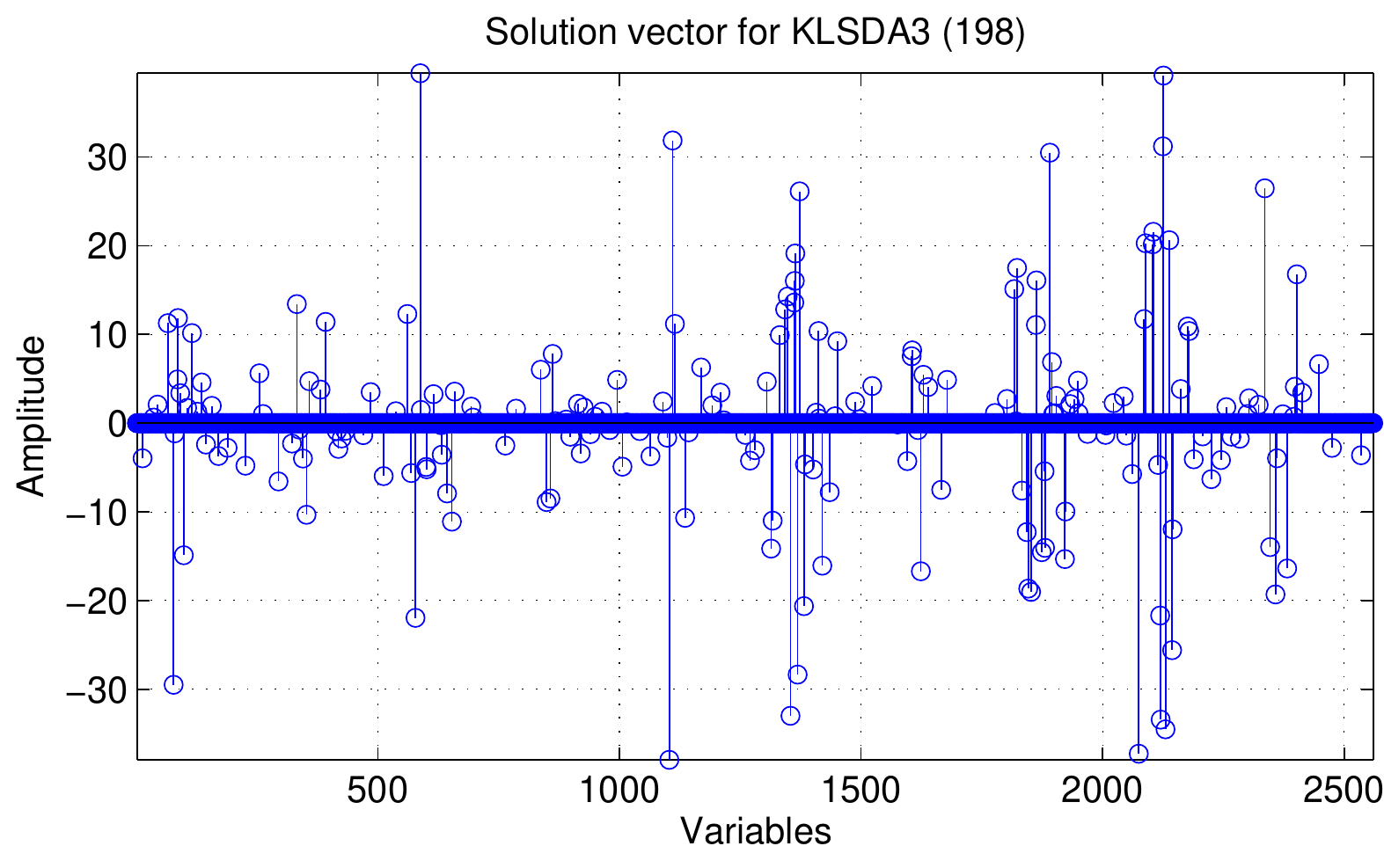}\\
\end{tabular}
\vspace{0.2in}
\caption{Solution vectors for one subject of Dataset-1 in each KLSDA configuration.}
\label{betas}
\end{figure}

Figure \ref{betas} depicts the morphology of the solution vectors for one subject of Dataset-1 in the four different KLSDA configurations. Between parentheses the number of non-zero values in the solution vector is shown. Note that this number is higher when the anisotropy matrix $\mathbf{D}$ is used in the $\ell_1$ penalization term (KLSDA1 and KLSDA3), and also the amplitudes of the coefficients increase in those two cases.  
%\begin{figure}
%\centerline{
%\includegraphics[width=0.9\textwidth]{betas}}
%\caption{Solution vectors for one subject of Dataset-1 in each KLSDA configuration}
%\label{betas}
%\end{figure}

\section{Discussion}
\label{Discussion}
It is well known that LDA is a commonly used method for ERP classification purposes. Its wide use is mainly due to its robustness, simplicity and good classification performances. When the number of observations is significantly lower than their dimensionality, LDA performs poorly, reason for which several alternatives to this method have been proposed. In the present work we described different LDA approaches from the statistical literature and developed a new penalized sparse discriminant analysis method called Kullback-Leibler Penalized Sparse Discriminant Analysis. This new method not only inherits the good properties of  SDA, but it also allows us to incorporate KLD-based discriminative information, in order to enhance classification performance.  

It is important to highlight that our implementation of KLSDA incorporates automatic tuning parameter selection. In light of the sparsity degree of the solution and the classification performances obtained, this procedure has proved to be very adequate for choosing the two regularization parameters in the model. 

We tested the new KLSDA approach with two real ERP-EEG datasets. An analysis of the classification results indicates that the KLSDA configuration leading to the best performance is subject depended. In particular, the classification results from Dataset-2 (ALS patients), show that adding KLD information into the solution may not necessarily result beneficial, specially when it provides no clear discriminative information. In regard to this, we shall compare these results with the ones obtained by using another discriminant information measure, as the asymmetric KLD or the ones presented in \cite{johnson2001symmetrizing, reid2011information}, to cite a few. 

It is remarkable that in those cases where KLSDA outperformed SDA there were no computational cost added. In fact, in our experiments we found that SDA's  computational cost was the same or even higher than KLSDA's with two anisotropic penalizing terms (KLSDA3). 

The results achieved by applying KLSDA in the context of ERP classification problems are high enough (over 75\%) to ensure good communication between the brain of the person and the device being controlled. These results encourage us to continue research efforts. There is clearly much room for improvement. Further research is currently underway in several directions. For instance, different discrepancy measures, anisotropy matrices and  penalizing terms can be considered.

A final remark is that, although the KLSDA method was inspired by the idea of solving the binary ERP classification problem in BCI systems, it can clearly be applied to any type of classification problem. 

\section*{Acknowledgements}
This work was supported in part by Consejo Nacional de Investigaciones Cient\'{\i}ficas y T\'{e}cnicas, CONICET, through PIP 2014-2016 No. 11220130100216-CO, the Air Force
Office of Scientific Research, AFOSR/SOARD, through Grant FA9550-14-1-0130 and and by Universidad Nacional del Litoral, UNL, through CAID-UNL 2011 Project No.525 within PACT ``Señales, Sistemas e Inteligencia Computacional''.




\section*{References}
\bibliographystyle{plain}
\bibliography{Biblio}

\begin{thebibliography}{10}

\bibitem{Distances}
M.~Basseville.
\newblock Distance measures for signal processing and pattern recognition.
\newblock {\em Signal Processing}, 18:349--369, 1989.

\bibitem{hyperlcurve}
Murat Belge, Misha~E Kilmer, and Eric~L Miller.
\newblock Simultaneous multiple regularization parameter selection by means of
  the {L}-hypersurface with applications to linear inverse problems posed in
  the wavelet transform domain.
\newblock In {\em SPIE's International Symposium on Optical Science,
  Engineering, and Instrumentation}, pages 328--336. International Society for
  Optics and Photonics, 1998.

\bibitem{significant}
Martin Billinger, Ian Daly, Vera Kaiser, Jing Jin, Brendan~Z Allison, Gernot~R
  M{\"u}ller-Putz, and Clemens Brunner.
\newblock Is it significant? guidelines for reporting {BCI} performance.
\newblock In {\em Towards Practical Brain-Computer Interfaces}, pages 333--354.
  Springer, 2012.

\bibitem{Blankertz-tutorial}
Benjamin Blankertz, Steven Lemm, Matthias Treder, Stefan Hauf, and Klaus~Robert
  M$\ddot{u}$ler.
\newblock Single-trial analysis and classification of {ERP} component- a
  tutorial.
\newblock {\em Neuroimage}, 56:814--825, 2011.

\bibitem{AUC}
Andrew~P Bradley.
\newblock The use of the area under the {ROC} curve in the evaluation of
  machine learning algorithms.
\newblock {\em Pattern recognition}, 30(7):1145--1159, 1997.

\bibitem{SLDA}
Line Clemmensen, Trevor Hastie, Daniela Witten, and Bjarne Ersb{\o}ll.
\newblock Sparse discriminant analysis.
\newblock {\em Technometrics}, 53:406--413, 2012.

\bibitem{Duda}
Richard~O Duda, Peter~E Hart, and David~G Stork.
\newblock {\em Pattern classification}.
\newblock John Wiley \& Sons, 2012.

\bibitem{Don-Far}
Lawrence~Ashley Farwell and Emanuel Donchin.
\newblock Talking off the top of your head: toward a mental prosthesis
  utilizing event-related brain potentials.
\newblock {\em Electroencephalography and clinical Neurophysiology},
  70(6):510--523, 1988.

\bibitem{ROC}
Tom Fawcett.
\newblock An introduction to {ROC} analysis.
\newblock {\em Pattern recognition letters}, 27(8):861--874, 2006.

\bibitem{fisher}
Ronald~A Fisher.
\newblock The use of multiple measurements in taxonomic problems.
\newblock {\em Annals of eugenics}, 7(2):179--188, 1936.

\bibitem{gareis2011}
Ivan~E Gareis, Ruben~C Acevedo, Yanina~V Atum, Gerardo~G Gentiletti,
  Veronica~Medina Banuelos, and Hugo~L Rufiner.
\newblock Determination of an optimal training strategy for a {BCI}
  classification task with {LDA}.
\newblock In {\em Neural Engineering (NER), 2011 5th International IEEE/EMBS
  Conference on}, pages 286--289. IEEE, 2011.

\bibitem{automatic}
Will Gersch, F~Martinelli, J~Yonemoto, MD~Low, and JA~Mc~Ewan.
\newblock Automatic classification of electroencephalograms: Kullback-leibler
  nearest neighbor rules.
\newblock {\em Science}, 205(4402):193--195, 1979.

\bibitem{KullbackClassification}
Anjum Gupta, Shibin Parameswaran, and Cheng-Han Lee.
\newblock Classification of electroencephalography ({EEG}) signals for
  different mental activities using kullback leibler (kl) divergence.
\newblock In {\em Acoustics, Speech and Signal Processing, 2009. ICASSP 2009.
  IEEE International Conference on}, pages 1697--1700. IEEE, 2009.

\bibitem{lcurve}
Per~Christian Hansen.
\newblock Analysis of discrete ill-posed problems by means of the {L}-curve.
\newblock {\em SIAM review}, 34(4):561--580, 1992.

\bibitem{penalized}
Trevor Hastie, Andreas Buja, and Robert Tibshirani.
\newblock Penalized discriminant analysis.
\newblock {\em The Annals of Statistics}, pages 73--102, 1995.

\bibitem{flexlda}
Trevor Hastie, Robert Tibshirani, and Andreas Buja.
\newblock Flexible discriminant analysis by optimal scoring.
\newblock {\em Journal of the American statistical association},
  89(428):1255--1270, 1994.

\bibitem{bookstats}
Trevor Hastie, Robert Tibshirani, and Jerome Friedman.
\newblock {\em The elements of statistical learning: data mining, inference and
  prediction}.
\newblock Springer Series in Statistics, 2009.

\bibitem{Electroph}
Steven~A Hillyard and Marta Kutas.
\newblock Electrophysiology of cognitive processing.
\newblock {\em Annual review of psychology}, 34(1):33--61, 1983.

\bibitem{meanshrinkage}
Johannes Hohne, Benjamin Blankertz, Klaus-Robert Muller, and Daniel Bartz.
\newblock Mean shrinkage improves the classification of {ERP} signals by
  exploiting additional label information.
\newblock In {\em Pattern Recognition in Neuroimaging, 2014 International
  Workshop on}, pages 1--4. IEEE, 2014.

\bibitem{johnson2001symmetrizing}
Don Johnson and Sinan Sinanovic.
\newblock Symmetrizing the kullback-leibler distance.
\newblock {\em IEEE Transactions on Information Theory}, 2001.

\bibitem{krusienskicomparison}
Dean~J Krusienski, Eric~W Sellers, Fran{\c{c}}ois Cabestaing, Sabri Bayoudh,
  Dennis~J McFarland, Theresa~M Vaughan, and Jonathan~R Wolpaw.
\newblock A comparison of classification techniques for the {P300} speller.
\newblock {\em Journal of neural engineering}, 3(4):299, 2006.

\bibitem{kullback}
Solomon Kullback and Richard~A Leibler.
\newblock On information and sufficiency.
\newblock {\em The annals of mathematical statistics}, 22(1):79--86, 1951.

\bibitem{base-datos}
Claudia Ledesma-Ramirez, Erik Bojorges-Valdez, Oscar Y{\'a}{\~n}ez-Suarez,
  Carolina Saavedra, Laurent Bougrain, and Gerardo~Gabriel Gentiletti.
\newblock An open-access {P300} speller database.
\newblock In {\em Fourth International Brain-Computer Interface Meeting}, 2010.

\bibitem{AdvanceChallenges}
K~Li, V~N Raju, R~Sankar, Y~Arbel, and E~Donchin.
\newblock Advances and challenges in signal analysis for single trial
  {P300-BCI}.
\newblock In {\em Foundations of Augmented Cognition. Directing the Future of
  Adaptive Systems}, volume~2, pages 87--94, ll, 2011.

\bibitem{Lottereview}
Fabien Lotte, Marco Congedo, Anatole L{\'e}cuyer, Fabrice Lamarche, and Bruno
  Arnaldi.
\newblock A review of classification algorithms for {EEG}-based brain-computer
  interfaces.
\newblock {\em Journal of neural engineering}, 4(2):R1, 2007.

\bibitem{mazzieri2015mixed}
Gisela~L Mazzieri, Ruben~D Spies, and Karina~G Temperini.
\newblock Mixed spatially varying {L2-BV} regularization of inverse ill-posed
  problems.
\newblock {\em Journal of Inverse and Ill-posed Problems}, 23(6):571--585,
  2015.

\bibitem{kullbackSVM}
Pedro~J Moreno, Purdy~P Ho, and Nuno Vasconcelos.
\newblock A kullback-leibler divergence based kernel for {SVM} classification
  in multimedia applications.
\newblock In {\em Advances in neural information processing systems}, page
  None, 2003.

\bibitem{Genet}
Geoffroy Mouret, Jean-Jules Brault, and Vahid Partovinia.
\newblock Generalized elastic net regression.
\newblock 2013.

\bibitem{reid2011information}
Mark~D Reid and Robert~C Williamson.
\newblock Information, divergence and risk for binary experiments.
\newblock {\em Journal of Machine Learning Research}, 12(Mar):731--817, 2011.

\bibitem{P300ALS}
Angela Riccio, Luca Simione, Francesca Schettini, Alessia Pizzimenti, Maurizio
  Inghilleri, Marta~Olivetti Belardinelli, Donatella Mattia, and Febo Cincotti.
\newblock Attention and {P300}-based {BCI} performance in people with
  amyotrophic lateral sclerosis.
\newblock {\em Front Hum Neurosci 2013 Nov; 7}, 732(2,906), 2013.

\bibitem{spasm}
Karl Sj{\"o}strand, Line~Harder Clemmensen, Rasmus Larsen, and Bjarne
  Ersb{\o}ll.
\newblock {SpaSM}: A matlab toolbox for sparse statistical modeling.
\newblock {\em Journal of Statistical Software Accepted for publication}, 2012.

\bibitem{lasso}
Robert Tibshirani.
\newblock Regression shrinkage and selection via the lasso.
\newblock {\em Journal of the Royal Statistical Society. Series B
  (Methodological)}, pages 267--288, 1996.

\bibitem{Glasso}
Ryan~Joseph Tibshirani and Jonathan~E Taylor.
\newblock The solution path of the generalized lasso.
\newblock {\em The Annals of Statistics}, 39:1335–1371, 2011.

\bibitem{wolpawBCI}
Jonathan Wolpaw and Elizabeth~Winter Wolpaw.
\newblock {\em Brain-Computer Interfaces: principles and practice}.
\newblock Oxford University Press, USA, 2012.

\bibitem{BCIcommunication}
Jonathan~R Wolpaw, Niels Birbaumer, Dennis~J McFarland, Gert Pfurtscheller, and
  Theresa~M Vaughan.
\newblock Brain-computer interfaces for communication and control.
\newblock {\em Clinical neurophysiology}, 113(6):767--791, 2002.

\bibitem{LSLDA}
Jieping Ye.
\newblock Leaste squares linear discriminant analysis.
\newblock In {\em Proceedings of the 24th international conference on Machine
  learning}, pages 1087--1093. ACM, 2007.

\bibitem{aggregation}
Yu~Zhang, Guoxu Zhou, Jing Jin, Qibin Zhao, Xingyu Wang, and Andrzej Cichocki.
\newblock Aggregation of sparse linear discriminant analyses for event-related
  potential classification in brain-computer interface.
\newblock {\em International journal of neural systems}, 24(01):1450003, 2014.

\bibitem{enet}
Hui Zou and Trevor Hastie.
\newblock Regularization and variable selection via the elastic net.
\newblock {\em Journal of the Royal Statistical Society: Series B (Statistical
  Methodology)}, 67(2):301--320, 2005.

\end{thebibliography}
%\begin{thebibliography}{00}
%%% Text of bibliographic item
%
%\bibitem{}
%
%\end{thebibliography}
\vspace{0.1in}
\end{document}